\documentclass{article}
\usepackage{arxiv}

\usepackage[utf8]{inputenc} 
\usepackage[T1]{fontenc}    
\usepackage{hyperref}       
\usepackage{url}            
\usepackage{booktabs}       
\usepackage{amsfonts}       
\usepackage{nicefrac}       
\usepackage{microtype}      
\usepackage{amsmath}
\title{Discriminative learning for probabilistic context-free grammars
  based on generalized H-criterion}
  \author{
  Mauricio Maca\\
  Department of Mathematics\\
  Universidad del Cauca\\
  \texttt{mmaca@unicauca.edu.co} \\
  \And
  Jos\'e Miguel Bened\'{i} and Joan Andreu S\'anchez \\
  PRHLT Research Center\\
  Universitat Polit\'ecnica de Val\'encia\\
  \texttt{\{jmbenedi,jandreu\}@prhlt.upv.es} \\
}
\begin{document}
\maketitle

\begin{abstract}
  We present a formal framework for the development of a family of
  discriminative learning algorithms for Probabilistic Context-Free
  Grammars (PCFGs) based on  a generalization of criterion-H.
  First of all, we propose the H-criterion as the objective function
  and the Growth Transformations as the optimization method, which
  allows us to develop the final expressions for the estimation of the
  parameters of the PCFGs.
  And second, we generalize the H-criterion to take into account the
  set of reference interpretations and the set of competing
  interpretations, and we propose a new family of objective functions
  that allow us to develop the expressions of the estimation
  transformations for PCFGs.
\end{abstract}

\section{Introduction}
\label{sec:intro}    

Throughout time, there has been an interest in Probabilistic
Context-Free Grammars (PCFGs) for use in different tasks within the
framework of Syntactic Pattern Recognition~\cite{Lari1991, Ney1992,
  Alvaro2016} and Computational Linguistics~\cite{Turian2006,
  Finkel2008}.
The reason for this can be found in the capability of PCFGs to model
the long-term dependencies established between the different
linguistic units of a sentence, and the possibility of incorporating
the probabilistic information which allows for adequate modeling of the
variability phenomena that are always present in complex problems.

Given a training sample, the problem of learning a PCFG can be stated
as an estimation process of the parameters of the PCFG.  To tackle
this estimation process two aspects have to be considered: proposing
an optimization method and defining a certain objective function. The
optimization method we have considered is based on the Growth
Transformation framework~\cite{Baum1968,Casacuberta1996a}, and the
classic objective function is based on the \textit{Maximum Likelihood
Estimation} (MLE) criterion.

Discriminative learning is a method that arises to improved
recognition accuracy for Natural Language Processing (NLP) problems
(\cite{Wang2003, He2008, Hsiao2009}). In this learning framework,
several objective functions were proposed: \textit{Maximum Mutual
  Information} (MMI) or \textit{Conditional Maximum Likelihood
  Estimation}, among others.

Some discriminative techniques for parsing (\cite{Taskar2004,
  Finkel2008}) get the features of their parsers of training set by
methods used in the unlexicalized generative parser for parsing
treebank. In general, only a parser for each sample is used so that
the resulting grammar is tractable.

Previous researches in Pattern Recognition have shown that parameters
estimation using discriminative techniques provides better performance
than the MLE training criterion. When MLE is used in parsing, the
parameters are reestimated to increase the likelihood of the parsers
of the training set without taking into count the probability of the
other possibles parsers. Whilst discriminative training techniques
consider possibles competing for the parser and reduce the probability
of incorrect parser.

In this work, we propose a discriminative method for learning PCFGs
based on a generalization of the H-criterion. Our formal framework
allows us simultaneously consider multiple reference trees.  We used
growth transformation as an optimization method to estimate parameters
and we noticed that it converges quickly. We build several
discriminative algorithms using the formal framework and they can be
implemented using well-known algorithms.

The paper is organized as follows. First, we present some related
works. Section 3 introduces the notation related to PCFGs. Section 4,
the H-criterion is presented as an objective function to solve the
estimation problem of PCFGs through the growth transformations
method. In Section 5, we propose a new generalization of the
H-criterion and we present some discriminative algorithms based on
this H-criterion. Finally, we present the conclusions and future work.

\section{Related work}
\label{sec:rw}    

Several parsers (\cite{Taskar2004, Finkel2008}), based on
discriminative training, use as input a generative component to
increase speed and accuracy. In (\cite{Taskar2004}), they employ for
the parser a max-margin principle of support vector machines.  They
transform the parsing problem to an optimization problem of a quadratic
program, which is solved through their dual problem. They train and
test on $\leq 15$ word sentences.

The parser presented in (\cite{Turian2006}) has no as input a
generative component as in (\cite{Taskar2004}). They use boost
decision trees to select compound features incrementally. For this,
the parser implements the search using an agenda that stores entire
states to build trees of decision. The parser improved training time
concerning the parser of(\cite{Taskar2004}), which uses five days
instead of several months.

In (\cite{Finkel2008}), they showed dynamic programming based
feature-rich discriminative parser. They defined a model based on
Conditional Random Fields used as a principal optimization method for
stochastic gradient descent.  They performed their experiments taking
into account two types of features: lexicon features, which are over
words and tags, and grammar features which obtained information of the
component generative.  The parser was trained and tested on sentences
of length $\leq 15$ and too was trained and tested on sentences of
length $\leq 40$.

\section{Preliminars}
\label{sec:pcfg}    

Before addressing the study of the discriminative estimation of PFCGs
using the H-criterion, we first introduce the notation about SCFGs
that is used in this work. 

A {\em Context-Free Grammar} (CFG) $G$ is a four-tuple
$(N,\Sigma,S,P)$, where $N$ is a finite set of non-terminals, $\Sigma$
is a finite set of terminals ($N \cap \Sigma = \emptyset$), $S \in N$
is the initial non-terminal, and $P$ is a finite set of rules: $A
\rightarrow \alpha$, $A \in N$, $\alpha \in (N \cup \Sigma)^+$ (we
only consider grammars with no empty rules). 
A CFG in Chomsky Normal Form (CNF) is a CFG in which the rules are of
the form $A \rightarrow BC$ or $A \rightarrow a$ ($A, B, C \in N$ and
$a \in \Sigma$).  
A {\em left-derivation} of $x \in \Sigma^+$ in $G$ is a sequence of
rules $d_x = (q_1, q_2, \ldots, q_m)$, $m \ge 1$, such that:
$(S \stackrel{q_1}{\Rightarrow} \alpha_1 \stackrel{q_2}{\Rightarrow}
\alpha_2 \stackrel{q_3}{\Rightarrow} \ldots
\stackrel{q_m}{\Rightarrow} x)$,
where $\alpha_i \in (N \cup \Sigma)^+$, $1 \leq i \leq m-1$ and $q_i$
rewrites the left-most non-terminal of $\alpha_{i-1}$. 
The {\em language generated} by $G$ is defined as
\(L(G)= \{ x \in \Sigma^+ \mid S \stackrel{*}{\Rightarrow} x\}\).
A CFG is called {\em ambiguous}, if for some $x \in L(G)$, there
exists more than one left-derivation.

\medskip
A {\em Probabilistic Context-Free Grammar} (PCFG) is defined as a pair
$G_p=(G,p)$, where $G$ is a CFG and $p:P \rightarrow ]0,1]$ is a
probability function of rule application such that
$\forall A \in N: \; \sum_{i=1}^{n_A} p(A \rightarrow \alpha_i) = 1;
\; \; $ where $n_A$ is the number of rules associated to~$A$.

Let $G_p$ be a PCFG. Then, for each $x \in L(G)$, we denote $D_x$ as
the set of all left-derivations of the string $x$. The expression
${\rm N}(A \rightarrow \alpha, d_x)$ represents the number of times
that the rule $A \rightarrow \alpha$ has been used in the derivation
$d_x$, and ${\rm N}(A, d_x)$ is the number of times that the
non-terminal $A$ has been derived in $d_x$. Obviously, this equation
is satisfied: $\; \; {\rm N}(A, d_x) = \sum_{i=1}^{n_A} {\rm N}(A
\rightarrow \alpha_i, d_x)$.

Then, we define the following expressions:

\begin{itemize}
\item {\em Probability of the derivation} $d_x$ of the string $x$
  as
\[
  {P}_{G_P}(x, d_x) =
  \prod_{\forall (A \rightarrow \alpha) \in P}
  p(A \rightarrow \alpha)^{{\rm N}(A \rightarrow \alpha, d_x)},
\]
\item {\em Probability} of the string $x$ as
\begin{equation}
  {P}_{G_p}(x) = \sum_{\forall d_x \in D_x} 
  {P}_{G_p}(x, d_x), \label{eq:f2} 
\end{equation}
\item {\em Probability of the best derivation} of the string
$x$ as
\begin{equation}
  \widehat{P}_{G_p}(x) = \max_{\forall d_x \in D_x}
  {P}_{G_p}(x, d_x),
  \label{eq:f3}
\end{equation}
\item {\em Best derivation} of the string $x$ as
\[ 
\widehat{d}_x = \arg\max_{\forall d_x \in D_x} {P}_{G_ps}(x, d_x). 
\]
\end{itemize}

The probability of the derivation $d_x$ of the string $x$ can be
interpreted as the joint probability of $x$ and $d_x$, so as the
probability of the string $x$ can be interpreted as the marginal
probability of $x$.

Given $\Delta_x \subseteq D_x$, a finite subset of derivations of $x$,
we can also define:

\begin{itemize}
\item Probability of $x$ with respect to $\Delta_x$,
\[
{P}_{G_p,\Delta_x}(x) = \sum_{\forall d_x \in \Delta_x}
{P}_{G_p}(x, d_x)
\]
\item Probability of the best derivation of $x$ with respect to
  $\Delta_x$
\[
\widehat{P}_{G_p, \Delta_x}(x) = \max_{\forall d_x \in \Delta_x} 
 {P}_{G_p}(x, d_x)
\]
\end{itemize}

These expressions respectively coincide with expression~\eqref{eq:f2}
and~\eqref{eq:f3} when $\Delta_x = D_x$.

Finally, the language generated by a PCFG $\; G_p$ is defined as:
$L(G_p) = \{x \in L(G) | {P}_{G_p}(x) > 0\}$. 
A PCFG $G_p$ is said to be {\em consistent} (\cite{Booth1973}) if the
language generated by $G_p$ is a probabilistic language, that is,
\( \sum_{x \in L(G_p)} {P}_{G_p}(x) = 1 \).



\bigskip %
Next we tackle the problem of estimating the parameters of a PCFG
$G_p = (G,p)$. This problem can be stated as follows: given a
probabilistic language $L_p = (L, \Phi)$ where $L$ is a language and
$\Phi$ is a probabilistic distribution over $L$, and given a training
sample $\Omega$, the estimation process consists in learning the
parameters of $G_p$ in order to represent $\Phi$ by means of the
probability~\eqref{eq:f2}.  Assuming that $\Omega$ is a representative
sample made up of a multi-set from $L$ according to $\Phi$, and
assuming that $\Phi$ can be represented by $G_p$, the estimation of
the parameters $p$ of $G_p$ is made by:
\[
\widehat{p} = \arg\max_p f_p(\Omega),  
\]
where $f_p(.)$ is a {\em objetive function} to be optimized. Two
issues have to be considered: the optimization method and the
selection of a objetive function.
In this paper we consider an optimization method, based on the {\em
  growth transformation} (GT)
framework~\cite{Baum1968,Casacuberta1996a}, and an objective function
derived from a generalization of the
H-criterion~\cite{Gopalakrishnan1988}.

\section{PCFGs estimation based on growth 
 transformations and H-criterion}
\label{sec:h-gt}    

In this section we use the restricted H-criteria as objective function
for discriminative training of a PCFG adapting it to the most recent
notation (\cite{Schluter2001, Woodland2002, He2008}).
We will first present the H-criterion and then we will develop the
method of growth transformations, applying the H-criterion, to
implement a discriminative learning method for estimating the
parameters of a PCFG.

\subsection{H-criterion}

The H-criterion based learning framework was proposed by
Gopalakrishnan, et al. in~\cite{Gopalakrishnan1988}, as a
generalization of the estimators of maximum likelihood (ML), maximum
mutual information (MMI) and conditional maximum likelihood (CML).
It is defined as follows: Let $a, b, c$ constants with $a > 0$. An
H-estimator $\widehat{\theta}(a, b, c)$ is obtained by minimizing the
H-criterion.

\vspace{-4mm}
\begin{align}
H_{a, b, c} (\theta; \Omega)\,=\,- \frac{1}{n}\;\sum_{i=1}^{n}\;\log\;
p_{\theta}^{a}(x_i, y_i)\;p_{\theta}^{b}(x_i)\;p_{\theta}^{c}(y_i)\;
\label{eq:h1}
\end{align}

Where $\Omega=\{(x_i, y_i)\}_{i=1}^N$ denotes the training sample,
$x_i$ are the observations and $y_i$ are the interpretations, and
$\theta$ are the parameters of the model. 

Thus the ML estimator is $\widehat{\theta} (1, 0, 0)$, the MMI
estimator is $\widehat{\theta} (1, -1, -1)$ and the CML estimator s
$\widehat{\theta} (1, 0, -1)$~\cite{Gopalakrishnan1988}.

\subsection{Objetive function based on H-criterion}

Given a PCFG, $G_p$, a training sample $\Omega$ and a set of
derivations $\Delta_x$, for each $x \in \Omega$, the estimation of the
probabilities of $G_p$ can be obtained through the H-criterion
minimizing the following estimator (see \eqref{eq:h1}),
\begin{align}
H_{1, -h, 0} (G_p, \Omega)\,=\,- \frac{1}{|\Omega|}\;\sum_{x\in \Omega}
   \;\log\;\frac{P_{G_p}(x, d_x)}{P_{G_p}(x)^h}\,=\, 
   - \frac{1}{|\Omega|}\;\log\;\prod_{x\in \Omega}\;
   \frac{P_{G_p}(x, d_x)}{P_{G_p}(x)^h} \label{eq:h2} 
\end{align}
where $0\leq h < 1$. In practice, the best derivation $\widehat{d_x}$
is used, and the total probability $P_{G_p}(x)$ is expanded by
marginalizing over all derivations of $x$ and maximizing the objective
function,
\begin{align}
{F_h} (G_p, \Omega)\,=\,\prod_{x\in \Omega}\;
   \frac{P_{G_p} (x, \widehat{d_x})} 
   {\left( \sum_{d_x\in \Delta_x}\;P_{G_p} (x, {d}_x)\right)^h}
   \,=\,\prod_{x\in \Omega}\;\frac{P_{G_p} (x, \widehat{d_x})} 
   {P_{G_p} (x, \Delta_x)^h}
\label{eq:h3}
\end{align}
The sum in the denominator of~\eqref{eq:h3} is the probability of
$x$ with respect to $\Delta_x$, where $\Delta_x$ denotes the set of
discriminated or competing derivations.  In this case, observations
are input strings, $x\in \Omega$, and interpretations are the
corresponding left-derivation sequences, $d_x$.
If $h > 0$ the H-criterion can be viewed as a discriminative training
method.  The exponent $h$ aims to establish the degree that competing
derivations discriminate against the derivation of reference. A
optimization of H-criterion attempts simultaneously to maximize the
numerator term $P_{G_p} (x, \widehat{d_x})$ and to minimize
denominator term $P_{G_p} (x, \Delta_x)^h$ for each string
$x\in \Omega$ in the training sample. 

Since ${F_h} (G_p, \Omega)$ is a rational function, the reduction of
the case of rational functions to polynomials proposed
in~\cite{Gopalakrishnan1991} can be applied.
\begin{align}
P_{\pi} (G_p, \Omega)\,=\, \prod_{x\in \Omega}\;P_{G_p} (x, \widehat{d_x})
  \,-\, \left(F_h(G_p, \Omega)\right)_{\pi}
  \;\prod_{x\in \Omega}\;P_{G_p} (x, \Delta_x)^h
\label{eq:h4}
\end{align}
Where $\left(F_h(G_p, \Omega)\right)_{\pi}$ is the constant that
results from evaluating $F_h(G_p, \Omega)$ at
$\pi$~\cite{Gopalakrishnan1991}.  $\pi$ is a point of the domain (in
our case $\pi$ will be the probabilities of the rules of $G_p$).

\subsection{Growth transformations for rational 
functions}

The objective function based on the H-criterion, and developed in
equations \eqref{eq:h3} and \eqref{eq:h4}, can be optimized by growth
transformations for rational functions~\cite{Gopalakrishnan1991}. And
the following final expression is obtained (see Appendix~A),
\begin{align}
&\bar{p}(A \rightarrow \alpha)\,=  \nonumber\\
&\qquad\qquad \frac{\sum_{x \in \Omega}\,\left[\,
  N(A \rightarrow \alpha, \widehat{d_{x}})\,-\,
  \frac{h}{P_{G_p}(x, \Delta_{x}})\,\sum_{d_{x} \in \Delta_{x}}\,
  N(A \rightarrow \alpha, d_{x})\,P_{G_p}(x, d_{x})\,\right]\,+\,
  p(A \rightarrow \alpha)\,\widetilde{C}}
  {\sum_{x \in \Omega}\, \left[\,N(A, \widehat{d_{x}})\,-\,
  \frac{h}{P_{G_p}(x, \Delta_{x})}\,\sum_{d_{x} \in \Delta_{x}}\,
  N(A, d_{x})\, P_{G_p}(x, d_{x})\,\right]\,+\,\widetilde{C}}
\label{eq:h5}
\end{align}

The value of the constant is
$C = \widetilde{C}\,\prod_{x \in \Omega}\;P_{G_p}(x,
\widehat{d_{x}})$.
Gopalakrishnan et al. suggested in~\cite{Gopalakrishnan1991} that to
obtain a fast convergence the constant $\widetilde{C}$ should be
calculated by means of the approximation,
\begin{align*}
\widetilde{C}\,=& \operatorname{max}\,\left\{\,
  \max_{p(A \rightarrow \alpha)}\,\left\{\,-\,
  \frac{1}{p(A \rightarrow \alpha)}\,\left[\,\sum_{x \in \Omega}\;
  N(A \rightarrow \alpha, \widehat{d_{x}})\right.\right.\right.\\
&\qquad\quad\left.\left.\left.\,-\,h\,\sum_{x \in \Omega}\;
  \frac{1}{P_{G_p}(x, \Delta_{x})}\;\sum_{d_{x} \in \Delta_{x}}\;
  N(A \rightarrow \alpha, d_{x})\,P_{G_p}(x, d_{x})\,
  \right]\,\right\}_{\pi}, \, 0\,\right\}\,+\,\epsilon
\end{align*}

where $\epsilon$ is a small positive constant. 

\medskip %
The growth transformation method allows us to easily obtain the
estimation of the probabilities of a PCFG
$\; \overline{G}_p = (G, \overline{p})$ from expression~\eqref{eq:h5}.
An iterative estimation process can be defined from
transformation~\eqref{eq:h5}. This process is carried out in two steps
on an initial PCFG until a local maximum is achieved.  In each
iteration, first the set $\Delta_x$ is computed for each
$x \in \Omega$ according to the selected criterion and then,
transformation~\eqref{eq:h5} is applied and a new PCFG is obtained
(more details can be found in Appendix~A).

\section{Generalization of the H-criterion for
   discriminative estimation of PCFGs}

In this section, we will explain the sense in which the H-criteria is
restricted, then we present the generalized H-criteria. Finally, we
find a growth transformation for generalized H-criteria and we define
some discriminative algorithm related.

\subsection{Generalized H-criterion}

Note that in expression~\eqref{eq:h3} we are considering that each
training sample has only a possible reference interpretation
(numerator in~\eqref{eq:h3}). Nevertheless, given the extremely
ambiguous nature of the models, the strings of training sample
presumably have more than one interpretation (parsing).
Therefore we will generalize the H-criteria to take into account this
event adapting it to unified criterion style introduced
by~\cite{Schluter2001}. We define the generalized H-criteria as:
\begin{align}
\widetilde{F}_h (G_p, \Omega)\,=\,\prod_{x\in \Omega}\;
     \frac{P_{G_p}^{\eta} (x, \Delta_x^{r})} 
   {P_{G_p} (x, \Delta_x^c)^h}
\label{eq:gh1}
\end{align}

where $0 < \eta, \; 0\leq h < 1\,$ and
$\Delta_x^{r} \subset \Delta_x^{c}$. The set $\Delta_x^{r}$ must
contain only derivations of correct parsing of the sentence $x$ while
the set $\Delta_x^{c}$ must contain competing derivations of any
parsing of the sentences.  Furthermore, it is satisfied that
$\widetilde{F_h} (G_p, \Omega) = \,{F_h} (G_p, \Omega)$ when
$\Delta_x^{r} = \{\widehat{d_x}\},\;\Delta_x^{c} = \Delta_x\,$ and
$\eta = 1$, therefore, it have the same set of maximum points.
Finally, if $h>0$ we can conclude that thee H-criterion can be viewed
as a discriminative training method.

\subsection{Growth transformations for generalized 
H-criterion}

The new objective function obtained from the generalization of the
H-criterion~\eqref{eq:gh1} can be optimized by means of growth
transformations for rational functions.  In a similar way to that used
in~\eqref{eq:h4}~\cite{Gopalakrishnan1991}, the rational function
$\widetilde{F}_h$ can be reduced to the polynomial function,
\begin{align}
Q_{\pi} (G_p, \Omega)\,=\, \prod_{x\in \Omega}\;P_{G_p}^{\eta} (x, \Delta_x^{r})
  \,-\, \left(\widetilde{F}_h(G_p, \Omega)\right)_{\pi}
  \;\prod_{x\in \Omega}\;P_{G_p} (x, \Delta_x^c)^h
\label{eq:gh2}
\end{align}
Where $\left(\widetilde{F}_h(G_p, \Omega)\right)_{\pi}$ is the
constant that results from evaluating $\widetilde{F}_h(G_p, \Omega)$
at $\pi$~\cite{Gopalakrishnan1991}. As in the previous case, $\pi$ is
a point of the domain (for us, $\pi$ will be the probabilities of the
rules of $G_p$).

\medskip %
As in the previous case, the new objective function obtained from the
generalization of the H-criterion~\eqref{eq:gh1} and~\eqref{eq:gh2}
can be optimized by growth transformations for rational
functions~\cite{Gopalakrishnan1991}. The complete development can be
found in Appendix B, and the final expression is as follows,
\begin{align}
\bar{p}(A \rightarrow \alpha)\,=\,
  \frac{{D}_{A \rightarrow \alpha} (\Delta_{x}^{r})\,-\,h\, 
  {D}_{A \rightarrow \alpha} (\Delta_{x}^{c})\,+\,
  p(A \rightarrow \alpha)\; \widetilde{C}}
  {{D}_{A} (\Delta_{x}^{r})\,-\,h\, {D}_{A} (\Delta_{x}^{c})\,+\,
  \widetilde{C}}
\label{eq:gh3}
\end{align}

where,
\begin{align*}
{D}_{A \rightarrow \alpha} (\Delta_{x}) &=\,\sum_{x \in \Omega}\;
   \frac{1}{{P}_{G_p}^{\eta} (x, \Delta_{x})}\, 
   \sum_{d_{x} \in \Delta_{x}}\;N(A \rightarrow \alpha, d_{x})\, 
   {P}_{G_p}^{\eta} (x, d_{x}) \\
{D}_{A} (\Delta_{x} &=\,\sum_{x \in \Omega}\;
   \frac{1}{{P}_{G_p}^{\eta} (x, \Delta_{x})}\,
   \sum_{d_{x} \in \Delta_{x}}\;N(A, d_{x})\,
   {P}_{G_p}^{\eta} (x, d_{x})
\end{align*}

The value of the constant is
$C = \widetilde{C}\;\eta\;\prod_{x \in \Omega}\;P_{G_p}^{\eta}(x,
\Delta_{x})$.
Following Gopalakrishnan et al. in~\cite{Gopalakrishnan1991}, and
carrying out a similar development to Section~\ref{sec:h-gt}, to
obtain a fast convergence the constant $\widetilde{C}$ should be
calculated by means of the approximation,
\begin{align*}
\widetilde{C}\,=& \operatorname{max}\,\left\{ \,
  \max_{p(A \rightarrow \alpha)}\,\left\{ \,-\,
  \frac{\left[\,{D}_{A \rightarrow \alpha}(\Delta_{x}^r)\,-\,h\,
   {D}_{A \rightarrow \alpha} (\Delta_{x}^c)\,\right]}
   {p(A \rightarrow \alpha)}\,
    \right\}_{\pi}, \, 0\,\right\}\,+\,\epsilon
\end{align*}

where $\epsilon$ is a small positive constant. 

\section{Discriminative algorithms based on 
  generalized H-criterion}

From transformation~\eqref{eq:gh3}, a family of discriminative
learning algorithms for PCFGs can be defined depending on how the set
of reference derivations $\Delta_x^r$ and the set of competing
derivations $\Delta_x^c$ are obtained, and the values of the
parameters $\eta$ and $h$.
For the algorithms studied here, we only analyze the effect of $h$
over the optimization framework and fix $\eta=1$.

The first issue to address is how the set of competing derivations can
be obtained.
If $\Delta_x^c$ is the set of all possible derivations, we can
calculate it using the well-known Inside algorithm~\cite{Lari1991}.
If $\Delta_x^c$ is the set of n-best derivations, we can calculate it
using a n-best parsing algorithm~\cite{Jimenez2000} and~\cite{Noya2020}.
Even,$\Delta_x^c$, may be the set of competing derivations that are
compaltible with a bracketed sample; in that case, we can use the
bracketed Inside algorithm~\cite{Pereira1992} and~\cite{Benedi2005}.

The second issue to consider is how the set of reference derivations
$\Delta_x^r$ is obtained. In any case, it must be satisfied that
$\Delta_x^r\, \subset \Delta_x^c$.
The set of reference derivations $\Delta_x^r$ may be the best
derivation, $\{\widehat{d_x}\}$, and can be calculated with the
well-known Viterbi algorithm~\cite{Ney1992}.
$\Delta_x^r$ may be the n-best derivations, and can be calculated with
the n-best parsing algorithm~\cite{Noya2020}.
Or $\Delta_x^r$ may be the best erivation that is compaltible with a
bracked sample, and can be calculated with the bracketed Viterbi
algorithm~\cite{Benedi2005}.

\subsection{Properties of the estimated models}

An important issue is the study of the PCFG's properties estimated by
discriminative algorithms based on the generalized H-criterion.  More
specifically, if the estimated PCFG generates a probabilistic
language, that is, if the estimated PCFG is consistent.  
Then we discuss the role of $h$ to ensure consistency of estimated
PCFGs.

As can be seen in~\eqref{eq:gh1}, if $h=0 (\eta=1)$ and
$\Delta_x^r=\{\widehat{d_x}\}$, the transformation~\eqref{eq:gh3} is
equivalent to the estimation algorithm based on the Viterbi Score
(VS)~\cite{Ney1992}. It is well known that PCFGs estimated by the VS
algorithm are always consistent~\cite{Sanchez1997}.

Next we explore what happens when $h=1$, and we show that we cannot
guarantee the consistency of the estimated models when $h=1$.  For
this, we consider that the reference derivations is selected as the
best derivation, $\Delta_x^r=\{\widehat{d_x}\}$, and the set of
competing derivations is the set of all possible derivation,
$\Delta_x^c=D_x$. And we illustrate this with an example:

\smallskip \hspace{5mm}\begin{minipage}{0.97\textwidth} 
  Let $G_p$ be an initial PCFG, where $N=\{ S \};\, \Sigma=\{ a\};\,$
  and $P=\{ (S \rightarrow S\,S, [q]),\, (S \rightarrow a, [1-q])\}$.
  We know that for values of $q$ that satisfy, $\,0.5 < q < 1$, the
  grammar $G_p$ is not consistent.

  \smallskip
  When training sample is $\{aa, aaaa\}$, there is only one derivation
  for aa with probability $q(1-p)^2$ and there are five derivations
  for aaaa each one of these with probability $q^3(1-q)^4$. Applying
  the transformation~\eqref{eq:gh3} we obtain,
  \begin{align*}
  p(S \rightarrow S\, S) &=\, 
    \frac{(1-h) + 3\,(1-h) + q\,C}{3\,(1-h) + 7\,(1-h) + C}\,=\,
    \frac{4\,(1-h) + q\,C}{10\,(1-h) + C}\,,
  \\
  p(S \rightarrow a) &=\, 
    \frac{2\,(1-h) + 4\,(1-h) + (1-q)\,C}
       {3\,(1-h) + 7\,(1-h) + C}\,=\,
    \frac{6\,(1-h) + (1-q)\,C}{10\,(1-h) + C}\,.
  \end{align*}

  If $h=1$ then $\bar{p}(S \rightarrow S\, S) = q$ and
  $\bar{p}(S \rightarrow a) = (1-q)$ preserving the inconsistency.
  However, if $0<h<1$, the consistency property is satisfied simply by
  setting $\epsilon = (1-h)$.
\end{minipage}

\section{Conclusion}

In this paper, we have presented a formal framework for the
development of a family of discriminative learning algorithms for
Probabilistic Context-Free Grammars (PCFGs) based on a generalization
of criterion-H.
First of all, we have presented the H-criterion as the objective
function and we have developed the final expressions for the
estimation of the parameters of the PCFGs.
Finally, we have proposed a generalization of the H-criterion to take
into account the set of reference interpretations and the set of
competing interpretations, and we have defined a new family of
objective functions that allow us to develop the expressions of the
estimation transformations for PCFGs.

\section*{Acknowledgment}

This work has been partially supported by the Ministerio de Ciencia y
Tecnología under the grant TIN2017-91452-EXP (IBEM).

\bibliographystyle{unsrt}  
\bibliography{references}

\section*{Appendix A}
\label{ap:A}

In this appendix, we demonstrate how expression~\eqref{eq:h5} is
derived in order to maximize expression~\eqref{eq:h4}, and then we
explain how the estimation process is carried out.  First, the growth
transformation optimization framework is presented, and
expression~\eqref{eq:h5} is formally derived applying the growth
transformations theorem for rational
functions~\cite{Gopalakrishnan1991} to optimize
expression~\eqref{eq:h4},
\begin{align}
\bar{p}(A \rightarrow \alpha)\,=\,\frac{p(A \rightarrow \alpha)\,
\left[\frac{\partial\, {P}_{\pi}(G_p, \Omega)}
  {\partial\, p(A \rightarrow \alpha)} + C\right]_{\pi}}
{\sum_{i=1}^{n_{A}}\; p(A \rightarrow \alpha_{i}) \left[
  \frac{\partial\, {P}_{\pi}(G_p, \Omega)}
  {\partial\, p(A \rightarrow \alpha_{i})} + C\right]_{\pi}}
\label{app:eq1}
\end{align}

where $n_A$ is the number of rules with the non-terminal $A$ in the
left side of the rule, and
$\pi = (\pi_{A_1}, \pi_{A_2}, \ldots, \pi_{A_{|N|}})$, $A_i \in N$,
$1 \leq i \leq |N|$ is a vector defined as follows:
$\pi_{A_i} = (p(A_i \rightarrow \alpha_{i1}), p(A_i \rightarrow
\alpha_{i2}), \ldots, p(A_i \rightarrow \alpha_{in_{A_i}}))$.
Furthermore, as shown in~\eqref{eq:h4}, ${P}_{\pi}(G_p, \Omega)$ is a
polynomial function where
$F_h(G_p,\Omega) = \prod_{x\in \Omega}\;\frac{P_{G_p} (x,
  \widehat{d_x})} {P_{G_p} (x, \Delta_x)^h}\,$
is a constant which is obtained by evaluating $F_h(G_p,\Omega)$ in
$\pi$. In~\cite{Gopalakrishnan1991} is shown that for every point of
the domain $\pi$ there is a constant $C$ such that the polynomial
$P_{\pi}+C$ has only non-negative coefficients.

Following a similar development to that used in~\cite{Benedi2005},
will allow us to explicitly obtain $\bar{p}(A \rightarrow \alpha)$
of~\eqref{app:eq1}.  The numerator of this expression can be written
as follows:
\begin{align*}
&p(A \rightarrow \alpha) \left(
  \frac{\partial\,\prod_{x \in \Omega} P_{G_p}(x, \widehat{d_x})}
  {\partial\, p(A \rightarrow \alpha)} - F_{\pi}(G_p, \Omega)\; 
  \frac{\partial\,\prod_{x \in \Omega} P_{G_p} (x, \Delta_{x})^{h}}
  {\partial\, p(A \rightarrow \alpha)} + C\right)_{\pi} \\
&\qquad=\, \left(\,\prod_{x \in \Omega} P_{G_p} (x, \widehat{d_{x}}) 
  \sum_{x \in \Omega}\; \frac{p(A \rightarrow \alpha)}
   {P_{G_p}(x, \widehat{d_{x}})}\;\frac{\partial\, 
    P_{G_p}(x, \widehat{d_{x}})}
    {\partial\, p(A \rightarrow \alpha)}\right.\\
&\qquad\qquad \left. - F_{\pi}(G_p, \Omega)\,\cdot\, h\, 
  \prod_{x \in \Omega} P_{G_p}(x, \Delta_{x})^{h}\, \sum_{x \in \Omega}\;
   \frac{p(A \rightarrow \alpha)}{P_{G_p}(x, \Delta_{x})}\; 
   \frac{\partial\, P_{G_p}(x, \Delta_{x})}
   {\partial\, p(A \rightarrow \alpha)}\ +\ p(A \rightarrow \alpha)\
   C\right)_{\pi} \\
&\qquad=\, \left(\,\prod_{x \in \Omega} P_{G_p} (x, \widehat{d_{x}})\,
  \left[\,\sum_{x \in \Omega}\; \frac{p(A \rightarrow \alpha)}
   {P_{G_p}(x, \widehat{d_{x}})}\;\frac{\partial\, 
    P_{G_p}(x, \widehat{d_{x}})}
    {\partial\, p(A \rightarrow \alpha)}\right.\right.\\
&\qquad\qquad \left.\left. -\,h\; \sum_{x \in \Omega}\;
   \frac{p(A \rightarrow \alpha)}{P_{G_p}(x, \Delta_{x})}\; 
   \frac{\partial\, P_{G_p}(x, \Delta_{x})}
   {\partial\, p(A \rightarrow \alpha)}\ \right]\ +\ 
   p(A \rightarrow \alpha)\ C\,\right)_{\pi} \\
&\qquad=\,\prod_{x \in \Omega} P_{G_p} (x, \widehat{d_{x}})\,\left[\;
  \sum_{x \in \Omega}\; N(A \rightarrow \alpha,\widehat{d_{x}})\right.\\ 
&\qquad\qquad \left. -\,h\; \sum_{x \in \Omega}\;
   \frac{1}{P_{G_p} (x, \Delta_{x})}\;\sum_{d_{x} \in \Delta_{x}}\;
   N(A \rightarrow \alpha, d_{x})\; P_{G_p}(x, d_{x})\,+\,
   \frac{p(A \rightarrow \alpha)\, C}
   {\prod_{x \in \Omega}\;P_{G_p}(x, \widehat{d_{x}})}\,\right]
\end{align*}

A similar expression can be obtained for the denominator:
\begin{align*}
&\sum_{i=1}^{n_{A}}\;p(A \rightarrow \alpha_i) \left[
  \frac{\partial\,\prod_{x \in \Omega} P_{G_p}(x, \widehat{d_x})}
  {\partial\, p(A \rightarrow \alpha_i)} - F_{\pi}(G_p, \Omega)\; 
  \frac{\partial\,\prod_{x \in \Omega} P_{G_p} (x, \Delta_{x})^{h}}
  {\partial\, p(A \rightarrow \alpha_i)} + C\right]_{\pi} \\
&\qquad=\,\sum_{i=1}^{n_{A}}\;\prod_{x \in \Omega} 
  P_{G_p}(x, \widehat{d_{x}})\,\left[\;\sum_{x \in \Omega}\; 
  N(A \rightarrow \alpha_i, \widehat{d_{x}})\right.\\
&\qquad\qquad \left. -\,h\; \sum_{x \in \Omega}\;
   \frac{1}{P_{G_p} (x, \Delta_{x})}\;\sum_{d_{x} \in \Delta_{x}}\;
   N(A \rightarrow \alpha_i, d_{x})\; P_{G_p}(x, d_{x})\,+\,
   \frac{p(A \rightarrow \alpha_i)\, C}
   {\prod_{x \in \Omega}\;P_{G_p}(x, \widehat{d_{x}})}\,\right] \\
&\qquad=\,\prod_{x \in \Omega}\;P_{G_p}(x, \widehat{d_{x}})\,
  \left[\;\sum_{x \in \Omega}\;\sum_{i=1}^{n_{A}}\;
  N(A \rightarrow \alpha_i, \widehat{d_{x}})\right.\\
&\qquad\qquad \left. -\,h\; \sum_{x \in \Omega}\;
   \frac{1}{P_{G_p} (x, \Delta_{x})}\;\sum_{d_{x} \in \Delta_{x}}\;
   \sum_{i=1}^{n_{A}}\;N(A \rightarrow \alpha_i, d_{x})\; 
   P_{G_p}(x, d_{x})\,+\,\frac{\sum_{i=1}^{n_{A}}\;
   p(A \rightarrow \alpha_i)\, C}
   {\prod_{x \in \Omega}\;P_{G_p}(x, \widehat{d_{x}})}\,\right] \\
&\qquad=\,\prod_{x \in \Omega}\;P_{G_p}(x, \widehat{d_{x}})\,
  \left[\;\sum_{x \in \Omega}\; 
  N(A, \widehat{d_{x}})\,-\,h\; \sum_{x \in \Omega}\;
   \frac{1}{P_{G_p} (x, \Delta_{x})}\;\sum_{d_{x} \in \Delta_{x}}\;
   N(A, d_{x})\;P_{G_p}(x, d_{x})\,+\,\frac{C}
   {\prod_{x \in \Omega}\;P_{G_p}(x, \widehat{d_{x}})}\,\right] \\
\end{align*}

Thus, expression~\eqref{app:eq1} can written as:
\begin{align*}
&\bar{p}(A \rightarrow \alpha)\,=  \nonumber\\
&\qquad\qquad \frac{\sum_{x \in \Omega}\,\left[\,
  N(A \rightarrow \alpha, \widehat{d_{x}})\,-\,
  \frac{h}{P_{G_p}(x, \Delta_{x}})\,\sum_{d_{x} \in \Delta_{x}}\,
  N(A \rightarrow \alpha, d_{x})\,P_{G_p}(x, d_{x})\,\right]\,+\,
  p(A \rightarrow \alpha)\,\widetilde{C}}
  {\sum_{x \in \Omega}\, \left[\,N(A, \widehat{d_{x}})\,-\,
  \frac{h}{P_{G_p}(x, \Delta_{x})}\,\sum_{d_{x} \in \Delta_{x}}\,
  N(A, d_{x})\, P_{G_p}(x, d_{x})\,\right]\,+\,\widetilde{C}}
\end{align*}
which coincides with expression~\eqref{eq:h5}.

\section*{Appendix B}
\label{ap:B}

In this appendix, we demonstrate how expression~\eqref{eq:gh3} is
derived in order to maximize expression~\eqref{eq:gh2}, and then we
explain how the estimation process is carried out. First, the growth
transformation is defined as,
\begin{align}
\bar{p}(A \rightarrow \alpha)\,=\,\frac{p(A \rightarrow \alpha)\,
\left[\frac{\partial\, {Q}_{\pi}(G_p, \Omega)}
  {\partial\, p(A \rightarrow \alpha)} + C\right]_{\pi}}
{\sum_{i=1}^{n_{A}}\; p(A \rightarrow \alpha_{i}) \left[
  \frac{\partial\, {Q}_{\pi}(G_p, \Omega)}
  {\partial\, p(A \rightarrow \alpha_{i})} + C\right]_{\pi}}
\label{apb:eq1}
\end{align}

where $n_A$ is the number of rules with the non-terminal $A$ in the
left side of the rule, and
$\pi = (\pi_{A_1}, \pi_{A_2}, \ldots, \pi_{A_{|N|}})$, $A_i \in N$,
$1 \leq i \leq |N|$ is a vector defined as follows:
$\pi_{A_i} = (p(A_i \rightarrow \alpha_{i1}), p(A_i \rightarrow
\alpha_{i2}), \ldots, p(A_i \rightarrow \alpha_{in_{A_i}}))$.
Furthermore, ${Q}_{\pi}(G_p, \Omega)$~\eqref{eq:gh2} is a polynomial
function and as demonstrated in~\cite{Gopalakrishnan1991}, for every
point of the domain $\pi$, there is a constant $C$ such that the
polynomial $P_{\pi}+C$ has only non-negative coefficients.
Following a similar development to that used in~\cite{Benedi2005},
will allow us to explicitly obtain $\bar{p}(A \rightarrow \alpha)$
of~\eqref{apb:eq1}.

Let's define an auxiliary function,
\begin{align*}
  \mathcal{D}_{A \rightarrow \alpha}^h (\Delta_x)\,=\, 
   p(A \rightarrow \alpha)\,
  \left[\,\frac{\partial\,\prod_{x\in \Omega}  
  P_{G_p}^{\eta} (x, \Delta_x)^h}
  {\partial\, p(A \rightarrow \alpha)}\,\right]_{\pi}
\end{align*}

then the expression~\eqref{apb:eq1} can be rewritten as, 
\begin{align}
\bar{p}(A \rightarrow \alpha)\,=\,
  \frac{\mathcal{D}_{A \rightarrow \alpha}^1\,(\Delta_x^r)\, -\ 
        \widetilde{F}_h(G_p, \Omega)\; 
        \mathcal{D}_{A \rightarrow \alpha}^h\,(\Delta_x^c)\, +\ 
        p(A \rightarrow \alpha)\; C}
  {\sum_{i=1}^{n_{A}}\;\mathcal{D}_{A \rightarrow \alpha_i}^1\,(\Delta_x^r)
        \, -\, \widetilde{F}_h(G_p, \Omega)\; \sum_{i=1}^{n_{A}}\;
        \mathcal{D}_{A \rightarrow \alpha_i}^h\,(\Delta_x^c)
        \ +\ p(A \rightarrow \alpha)\;C}
\label{apb:eq3}
\end{align}

First, the auxiliary function
$\mathcal{D}_{A \rightarrow \alpha}^h (\Delta_x)$ is computed to
evaluate $\mathcal{D}_{A \rightarrow \alpha}^1 (\Delta_x^r)$ and
$\mathcal{D}_{A \rightarrow \alpha}^h (\Delta_x^c)$ in the numerator.
\begin{align}
&\mathcal{D}_{A \rightarrow \alpha}^{h} (\Delta_{x})\,=\nonumber\\
&\qquad=\, p(A \rightarrow \alpha)\,\left[\,h\;\prod_{x \in \Omega}
  {P}_{G_p}^{\eta} (x, \Delta_{x})^{h-1}\;
  \frac{\partial \prod_{x \in \Omega}\,{P}_{G_p}^{\eta} (x, \Delta_{x})}
  {\partial\, p(A \rightarrow \alpha)}\,\right]_{\pi} \nonumber\\
&\qquad=\,h\,\left[\,\prod_{x \in \Omega}\,
  {P}_{G_p}^{\eta} (x, \Delta_{x})^{h}\;\sum_{x \in \Omega}\, 
  \frac{p(A \rightarrow \alpha)}{{P}_{G_p}^{\eta} (x, \Delta_{x})}\,
  \frac{\partial {P}_{G_p}^{\eta} (x, \Delta_{x})}
  {\partial\, p(A \rightarrow \alpha)}\,\right]_{\pi} \nonumber\\
&\qquad=\,h\,\eta\,\left[\,\prod_{x \in \Omega}\,
  {P}_{G_p}^{\eta} (x, \Delta_{x})^{h}\;\sum_{x \in \Omega}\, 
  \frac{1}{{P}_{G_p}^{\eta} (x, \Delta_{x})}\,
  \sum_{d_{x} \in \Delta_{x}}\,N(A \rightarrow \alpha, d_{x})\,
  {P}_{G_p}^{\eta} (x, d_{x})\,\right]_{\pi} 
\label{apb:eq4}
\end{align}

Then, the expression
$\sum_{i=1}^{n_{A}}\,\mathcal{D}_{A \rightarrow \alpha_i}^h
(\Delta_x)$
is computed to evaluate
$\sum_{i=1}^{n_{A}}\,\mathcal{D}_{A \rightarrow \alpha_i}^1
(\Delta_x^r)$
and
$\sum_{i=1}^{n_{A}}\,\mathcal{D}_{A \rightarrow \alpha_i}^h
(\Delta_x^c)$ in the denominator.
\begin{align}
&\sum_{i=1}^{n_{A}}\;
  \mathcal{D}_{A \rightarrow \alpha_{i}}^{h} (\Delta_{x})\,=\nonumber\\
&\qquad=\,\sum_{i=1}^{n_{A}}\;h\,\eta\,\left[\,\prod_{x \in \Omega}\,
  {P}_{G_p}^{\eta} (x, \Delta_{x})^{h}\;\sum_{x \in \Omega}\, 
  \frac{1}{{P}_{G_p}^{\eta} (x, \Delta_{x})}\,
  \sum_{d_{x} \in \Delta_{x}}\,N(A \rightarrow \alpha_i, d_{x})\,
  {P}_{G_p}^{\eta} (x, d_{x})\,\right]_{\pi}\nonumber\\ 
&\qquad=\,h\,\eta\,\left[\,\prod_{x \in \Omega}\,
  {P}_{G_p}^{\eta} (x, \Delta_{x})^{h}\;\sum_{x \in \Omega}\, 
  \frac{1}{{P}_{G_p}^{\eta} (x, \Delta_{x})}\,
  \sum_{d_{x} \in \Delta_{x}}\;\sum_{i=1}^{n_{A}}\;
  N(A \rightarrow \alpha_i, d_{x})\,
  {P}_{G_p}^{\eta} (x, d_{x})\,\right]_{\pi}\nonumber\\ 
&\qquad=\,h\,\eta\,\left[\,\prod_{x \in \Omega}\,
  {P}_{G_p}^{\eta} (x, \Delta_{x})^{h}\;\sum_{x \in \Omega}\, 
  \frac{1}{{P}_{G_p}^{\eta} (x, \Delta_{x})}\,
  \sum_{d_{x} \in \Delta_{x}}\;N(A, d_{x})\,
  {P}_{G_p}^{\eta} (x, d_{x})\,\right]_{\pi}
\label{apb:eq5}
\end{align}

Substituting the expressions~\eqref{apb:eq4} and~\eqref{apb:eq5} in
the transformation~\eqref{apb:eq3} and simplifying
$\eta\,\prod_{x \in \Omega}\,{P}_{G_p}^{\eta} (x, \Delta_{x})$ in
the numerator and denominator, results in:
\begin{align*}
\bar{p}(A \rightarrow \alpha)\,=\,
  \frac{{D}_{A \rightarrow \alpha} (\Delta_{x}^{r})\,-\,h\, 
  {D}_{A \rightarrow \alpha} (\Delta_{x}^{c})\,+\,
  p(A \rightarrow \alpha) \widetilde{C}}
  {{D}_{A} (\Delta_{x}^{r})\,-\,h\, {D}_{A} (\Delta_{x}^{c})\,+\,
  \widetilde{C}}
\end{align*}

which coincides with expression~\eqref{eq:gh3}. Where,
\begin{align*}
{D}_{A \rightarrow \alpha} (\Delta_{x}) &=\,\sum_{x \in \Omega}\;
   \frac{1}{{P}_{G_p}^{\eta} (x, \Delta_{x})}\, 
   \sum_{d_{x} \in \Delta_{x}}\;N(A \rightarrow \alpha, d_{x})\, 
   {P}_{G_p}^{\eta} (x, d_{x}) \\
{D}_{A} (\Delta_{x} &=\,\sum_{x \in \Omega}\;
   \frac{1}{{P}_{G_p}^{\eta} (x, \Delta_{x})}\,
   \sum_{d_{x} \in \Delta_{x}}\;N(A, d_{x})\,
   {P}_{G_p}^{\eta} (x, d_{x})
\end{align*}

\end{document}